\documentclass{article}
\usepackage{amssymb}
\usepackage{multirow}
\usepackage{synttree}
\usepackage{lineno,hyperref}
\usepackage{natbib}

\begin{document}

\title{Robust Multilingual Named Entity Recognition with Shallow Semi-Supervised Features\thanks{Please cite this paper as: R. Agerri, G. Rigau. Robust multilingual Named Entity Recognition with shallow semi-supervised features, Artificial Intelligence (2016). \url{http://dx.doi.org/10.1016/j.artint.2016.05.003}. \copyright  2016.}  \thanks{This manuscript version is made available under the CC-BY-NC-ND 4.0 license \url{http://creativecommons.org/licenses/by-nc-nd/4.0/}. Paper submitted 1 September 2015, Revised 11 May 2016, Accepted 15 May 2016.}}

\author{Rodrigo Agerri\thanks{Corresponding author: \texttt{rodrigo.agerri@ehu.eus}} and German Rigau\\
IXA NLP Group\\University of the Basque Country (UPV/EHU)\\
Donostia-San Sebasti\'an, Spain}
\date{}
\maketitle

\begin{abstract}
We present a multilingual Named Entity Recognition approach based on a robust and general set of features
across languages and datasets. Our system combines shallow local information with clustering
semi-supervised features induced on large amounts of unlabeled text. Understanding via empirical
experimentation how to effectively \emph{combine} various types of
\emph{clustering features} allows us to seamlessly export our system to other datasets and languages. The result is a simple but highly competitive system
which obtains state of the art results across five languages and twelve datasets.
The results are reported on standard shared task evaluation data such as CoNLL for English, Spanish
and Dutch. Furthermore, and despite the lack of linguistically motivated features, we also report
best results for languages such as Basque and German. In addition, we
demonstrate that our method also obtains very competitive results even when the amount of supervised
data is cut by half, alleviating the dependency on manually annotated data. Finally, the results show that our emphasis on
clustering features is crucial to develop robust out-of-domain models. The system
and models are freely available to facilitate its use and guarantee the reproducibility
of results.
\end{abstract}


\section{Introduction}\label{sec:introduction}

A named entity can be mentioned using a great variety of surface forms (Barack Obama, President Obama, Mr. Obama, B. Obama, etc.) and the same surface form can refer to a variety of named entities. For example, according to the English Wikipedia, the form `Europe' can ambiguously be used to refer to 18 different entities, including the continent, the European Union, various Greek mythological entities, a rock band, some music albums, a magazine, a short story, etc.\footnote{\url{http://en.wikipedia.org/wiki/Europe_(disambiguation)}} Furthermore, it is possible to refer to a named entity by means of anaphoric pronouns and co-referent expressions such as `he', `her', `their', `I', `the 35 year old', etc. Therefore, in order to provide an adequate and comprehensive account of named entities in text it is necessary to \emph{recognize} the mention of a named entity and to \emph{classify} it by a pre-defined type (e.g, person, location, organization). Named Entity Recognition and Classification (NERC) is usually a required step to perform Named Entity Disambiguation (NED), namely to link `Europe' to the right Wikipedia article, and to resolve every form of mentioning or co-referring to the same entity.

Nowadays NERC systems are widely being used in research for tasks such as Coreference Resolution \citep{pradhan-etal-conll-st-2012-ontonotes}, Named Entity Disambiguation \citep{cucerzan_large-scale_2007,han_generative_2011,hoffart_robust_2011,mendes_evaluating_2011,hachey_evaluating_2012} for which a lot of interest has been created by the TAC KBP shared tasks \citep{ji2011knowledge}, Machine Translation \citep{al2002translating,koehn_moses:_2007,babych2003improving,li2013name}, Aspect Based Sentiment Analysis \citep{liu_sentiment_2012,cambria2013new,pontiki2014semeval,pontiki-EtAl:2015:SemEval}, Event Extraction \citep{doddington2004automatic,ahn2006stages,ji2008refining,cybulska-vossen:2013:RANLP-2013,hong_using_2011} and Event Ordering \citep{minard-EtAl:2015:SemEval}.

Moreover, NERC systems are integrated in the processing chain of many industrial software applications, mostly by companies offering specific solutions for a particular industrial sector which require recognizing named entities specific of their domain. There is therefore a clear interest in both academic research and industry to develop robust and efficient NERC systems: For industrial vendors it is particularly important to diversify their services by including NLP technology for a variety of languages whereas in academic research NERC is one of the foundations of many other NLP end-tasks.

Most NERC taggers are supervised statistical systems that extract patterns and term features which are considered to be indications of Named Entity (NE) types using the manually annotated training data (extracting orthographic, linguistic and other types of evidence) and often external knowledge resources. As in other NLP tasks, supervised statistical NERC systems are more robust and obtain better performance on available evaluation sets, although sometimes the statistical models can also be combined with specific rules for some NE types. For best performance, supervised statistical approaches require manually annotated training data, which is both expensive and time-consuming. This has seriously hindered the development of robust high performing NERC systems for many languages but also for other domains and text genres \citep{nobata2000comparison,ritter_named_2011}, in what we will henceforth call `out-of-domain' evaluations.

Moreover, supervised NERC systems often require fine-tuning for each language and, as some of the features require language-specific knowledge, this poses yet an extra complication for the development of robust multilingual NERC systems. For example, it is well-known that in German every noun is capitalized and that compounds including named entities are pervasive. This also applies to agglutinative languages such as Basque, Korean, Finnish, Japanese, Hungarian or Turkish. For this type of languages, it had usually been assumed that linguistic features (typically Part of Speech (POS) and lemmas, but also semantic features based on WordNet, for example) and perhaps specific hand-crafted rules, were a necessary condition for good NERC performance as they would allow to capture better the most recurrent declensions (cases) of named entities for Basque \citep{alegria2006lessons} or to address problems such as sparsity and capitalization of every noun for German \citep{faruqui_training_2010,benikova7germeval,germaner}. This language dependency was easy to see in the CoNLL 2002 and 2003 tasks, in which systems participating in the two available languages for each edition obtained in general different results for each language. This suggests that without fine-tuning for each corpus and language, the systems did not generalize well across languages \citep{nothman_learning_2012}.

This paper presents a multilingual and robust NERC system based on simple, general and shallow features that heavily relies on word representation features for high performance. Even though we do not use linguistic motivated features, our approach also works well for inflected languages such as Basque and German. We demonstrate the robustness of our approach by reporting best results for five languages (Basque, Dutch, German, English and Spanish) on 12 different datasets, including seven in-domain and eight out-of-domain evaluations.

\subsection{Contributions}\label{sec:contributions}

The main contributions of this paper are the following: First, we show how to easily develop robust NERC systems across datasets and languages with minimal human intervention, even for languages with declension and/or complex morphology. Second, we empirically show how to effectively use various types of simple word representation features thereby providing a clear methodology for choosing and combining them. Third, we demonstrate that our system still obtains very competitive results even when the supervised data is reduced by half (even less in some cases), alleviating the dependency on costly hand annotated data. These three main contributions are based on:

\begin{enumerate}
\item A simple and shallow robust set of features across languages and datasets, even in out-of-domain evaluations.
\item The lack of linguistic motivated features, even for languages with agglutinative (e.g., Basque) and/or complex morphology (e.g., German).
\item A clear methodology for using and combining various types of word representation features by leveraging public unlabeled data.
\end{enumerate}

Our approach consists of shallow local features complemented by three types of word representation (clustering) features: Brown clusters \citep{brown1992class}, Clark clusters \citep{clark2003combining} and K-means clusters on top of the word vectors obtained by using the Skip-gram algorithm \citep{mikolov2013distributed}. We demonstrate that \emph{combining} and \emph{stacking} different clustering features induced from various data sources (Reuters, Wikipedia, Gigaword, etc.) allows to cover different and more varied types of named entities without manual feature tuning. Even though our approach is much simpler than most, we obtain the best results for Dutch, Spanish and English and comparable results in German (on CoNLL 2002 and 2003). We also report best results for German using the GermEval 2014 shared task data and for Basque using the Egunkaria testset \citep{alegria2006lessons}.

We report out-of-domain evaluations in three languages (Dutch, English and Spanish) using four different datasets to compare our system with the best publicly available systems for those languages: Illinois NER \citep{ratinov_design_2009} for English, Stanford NER \citep{finkel_incorporating_2005} for English and Spanish, SONAR-1 NERD for Dutch \citep{desmet2014fine} and Freeling for Spanish \citep{freeling3_padro12}.  We outperform every other system in the eight out-of-domain evaluations reported in Section \ref{sec:out-doma-eval}. Furthermore, the out-of-domain results show that our clustering features provide a simple and easy method to improve the robustness of NERC systems.

Finally, and inspired by previous work \citep{koo-carreras-collins:2008:ACLMain,biemann2009unsupervised} we measure how much supervision is required to obtain state of the art results. In Section \ref{sec:reduc-train-data} we show that we can still obtain very competitive results reducing the supervised data by half (and sometimes even more). This, together with the lack of linguistic features, means that our system considerably saves data annotation costs, which is quite convenient when trying to develop a NERC system for a new language and/or domain.

Our system learns Perceptron models \citep{collins_discriminative_2002} using the Machine Learning machinery provided by the Apache OpenNLP project\footnote{\url{http://opennlp.apache.org/}} with our own customized (local and clustering) features. Our NERC system is publicly available and distributed under the Apache 2.0 License and part of the IXA pipes tools \citep{AGERRI14.775.L14-1605}. Every result reported in this paper is obtained using the conlleval script from the CoNLL 2002 and CoNLL 2003 shared tasks\footnote{\url{http://www.cnts.ua.ac.be/conll2002/ner/bin/conlleval.txt}}. To guarantee reproducibility of results we also make publicly available the models and the scripts used to perform the evaluations. The system, models and evaluation scripts can be found in the \textit{ixa-pipe-nerc} website\footnote{\url{https://github.com/ixa-ehu/ixa-pipe-nerc}}.

Next Section reviews related work, focusing on best performing NERC systems for each language evaluated on standard shared evaluation task data. Section \ref{sec:system-description} presents the design of our system and our overall approach to NERC. In Section \ref{sec:results} we report the evaluation results obtained by our system for 5 languages (Basque, Dutch, German, English and Spanish) on 12 different datasets, distributed in 7 in-domain and 8 out-of-domain evaluations. Section \ref{sec:disc-future-work} discusses the results and contributions of our approach. In Section \ref{sec:conclusion} we highlight the main aspects of our work providing some concluding remarks and future work to be done using our NERC approach applied to other text genres, domains and sequence labeling tasks.

\section{Related Work}\label{sec:background}

The Named Entity Recognition and Classification (NERC) task was first defined for the Sixth Message Understanding Conference (MUC 6) \citep{nadeau_survey_2007}. The MUC 6 tasks focused on Information Extraction (IE) from unstructured text and NERC was deemed to be an important IE sub-task with the aim of recognizing and classifying nominal mentions of persons, organizations and locations, and also numeric expressions of dates, money, percentage and time. In the following years, research on NERC increased as it was considered to be a crucial source of information for other Natural Language Processing tasks such as Question Answering (QA) and Textual Entailment (RTE) \citep{nadeau_survey_2007}. Furthermore, while MUC 6 was solely devoted to English as target language, the CoNLL shared tasks (2002 and 2003) boosted research on language independent NERC for 3 additional target languages: Dutch, German and Spanish \citep{tjong_kim_sang_introduction_2002,tjong_kim_sang_introduction_2003}.

The various MUC, ACE and CoNLL evaluations provided a very convenient framework to test and compare NERC systems, algorithms and approaches. They provided manually annotated data for training and testing the systems as well as an objective evaluation methodology. Using such framework, research rapidly evolved from rule-based approaches (consisting of manually handcrafted rules) to language independent systems focused on learning supervised statistical models. Thus, while in the MUC 6 competition 5 out of 8 systems were rule-based, in CoNLL 2003 16 teams participated in the English task all using statistical-based NERC \citep{nadeau_survey_2007}.

\subsection{Datasets}\label{sec:datasets}

Table \ref{tab:datasets} describes the 12 datasets used in this paper. The first half lists the corpora used for in-domain evaluation whereas the lower half contains the out-of-domain datasets. The CoNLL NER shared tasks focused on language independent machine learning approaches for 4 entity types: \emph{person}, \emph{location}, \emph{organization} and \emph{miscellaneous} entities. The 2002 edition provided manually annotated data in Dutch and Spanish whereas in 2003 the languages were German and English. In addition to the CoNLL data, for English we also use the formal run of MUC 7 and Wikigold for out-of-domain evaluation. Very detailed descriptions of CoNLL and MUC data can easily be found in the literature, including the shared task descriptions themselves \citep{chinchor_muc-7_1998,tjong_kim_sang_introduction_2002,tjong_kim_sang_introduction_2003}, so in the following we will describe the remaining, newer datasets.

\begin{table*}[ht]\footnotesize
  \centering
  \begin{tabular}{cllrrrrrr} \hline
   & Corpus & Source & \multicolumn{6}{c}{Number of Tokens and Named Entities} \\ \hline
    & & & \multicolumn{2}{c}{train} & \multicolumn{2}{c}{dev} & \multicolumn{2}{c}{test} \\ \cline{4-9}
    & & & tok & ne & tok & ne & tok & ne \\ \hline \hline
    & \multicolumn{2}{l}{In-domain datasets}\\ \hline
    en & CoNLL 2003 & Reuters RCV1 & 203621 & 23499 & 51362 & 5942 & 46435 & 5648 \\
    de & CoNLL 2003 & Frankfurter Rundschau 1992 & 206931 & 11851 & 51444 & 4833 & 51943 & 3673 \\
    & GermEval 2014 & Wikipedia/LCC news & 452853 & 31545 & 41653 & 2886 & 96499 & 6893 \\
    es & CoNLL 2002 & EFE 2000 & 264715 & 18798 & 52923 & 4352 & 51533 & 3558 \\
    nl & CoNLL 2002 & De Morgen 2000 & 199069 & 13344 & 36908 & 2616 & 67473 & 3941 \\
    eu & Egunkaria & Egunkaria 1999-2003 & 44408 & 3817 & & & 15351 & 931 \\ \hline
    & \multicolumn{2}{l}{Out-of-domain datasets}\\ \hline
    en & MUC7 & newswire & & & & & 53749 & 3514 \\
    & Wikigold & Wikipedia 2008 & & & & & 39007 & 3558 \\
    & MEANTIME & Wikinews 2013 & & & & & 13957 & 1432 \\
    nl & SONAR-1 & various genres & & & & & 1000000 & 62505 \\
    & MEANTIME & Wikinews 2013 & & & & & 13425 & 1545 \\
    es & Ancora 2.0 & newswire & 547198 & 36938 \\
    & MEANTIME & Wikinews 2013 & 15853 & 1706 \\ \hline
  \end{tabular}
  \caption{Datasets used for training, development and evaluation. MUC7: only three classes (LOC, ORG, PER) of the formal run are used for out-of-domain evaluation. As there are not standard partitions of SONAR-1 and Ancora 2.0, the full corpus was used for training and later evaluated in-out-of-domain settings.}
  \label{tab:datasets}
\end{table*}

The Wikigold corpus consists of 39K words of English Wikipedia manually annotated following the CoNLL 2003 guidelines \citep{nothman_learning_2012}. For Spanish and Dutch, we also use Ancora 2.0 \citep{taule2008ancora} and SONAR-1 \citep{desmet2014fine} respectively. SONAR-1 is a one million word Dutch corpus with both coarse-grained and fine-grained named entity annotations. The coarse-grained level includes \emph{product} and \emph{event} entity types in addition to the four types defined in CoNLL data. Ancora adds \emph{date} and \emph{number} types to the CoNLL four main types. In Basque the only gold standard corpus is Egunkaria \citep{alegria2006lessons}. Although the Basque Egunkaria dataset is annotated with four entity types, the \emph{miscellaneous} class is extremely sparse, occurring only in a proportion of 1 to 10. Thus, in the training data there are 156 entities annotated as MISC whereas each of the other three classes contain around 1200 entities.

In the datasets described so far, named entities were assumed to be non-recursive and non-overlapping. During the annotation process, if a named entity was embedded in a longer one, then only the longest mention was annotated. The exceptions are the GermEval 2014 shared task data for German and MEANTIME, where nested entities are also annotated (both inner and outer spans).

The GermEval 2014 NER shared task \citep{benikova7germeval} aimed at improving the state of the art of German NERC which was perceived to be comparatively lower than the English NERC. Two main extensions were introduced in GermEval 2014; (i) fine grained named entity sub-types to indicate \emph{derivations} and \emph{compounds}; (ii) embedded entities (and not only the longest span) are annotated. In total, there are 12 types for classification: \emph{person}, \textit{location}, \textit{organization}, \textit{other} plus their sub-types annotated at their inner and outer levels.

Finally, the MEANTIME corpus \citep{MEANTIME:2016} is a multilingual (Dutch, English, Italian and Spanish) publicly available evaluation set annotated within the Newsreader project\footnote{\url{http://www.newsreader-project.eu}}. It consists of 120 documents, divided into 4 topics: Apple Inc., Airbus and Boeing, General Motors, Chrysler and Ford, and the stock market. The articles are selected in such a way that the corpus contains different articles that deal with the same topic over time (e.g. launch of a new product, discussion of the same financial indexes). Moreover, it contains nested entities so the evaluation results will be provided in terms of the outer and the inner spans of the named entities. MEANTIME includes six named entity types: \emph{person}, \emph{location}, \emph{organization}, \emph{product}, \emph{financial} and \emph{mixed}.

\subsection{Related Approaches}\label{sec:related-approaches}

Named entity recognition is a task with a long history in NLP. Therefore, we will summarize those approaches that are most relevant to our work, especially those we will directly compared with in Section \ref{sec:results}. Since CoNLL shared tasks, the most competitive approaches have been supervised systems learning CRF, SVM, Maximum Entropy or Averaged Perceptron models. In any case, while the machine learning method is important, it has also been demonstrated that good performance might largely be due to the feature set used \citep{ClarkCurran:2003}. Table \ref{tab:features} provides an overview of the features used by previous best scoring approaches for each of the five languages we address in this paper.

\begin{table*}[ht]\footnotesize
  \centering
  \begin{tabular}{lccccccccc} \hline
   System & Local & Ling & Global & Gaz & WR & Rules & Ensemble & Public & Res \\ \hline \hline
   Ratinov and Roth 2009 & \checkmark &  & \checkmark & \checkmark & \checkmark & & & \checkmark & \checkmark \\
   Passos et al. 2014 & \checkmark &  &  & \checkmark & \checkmark & & \checkmark \\
   ExB & \checkmark & \checkmark &  & \checkmark & \checkmark & \checkmark & \checkmark \\
   Faruqui et al. 2010 & \checkmark & \checkmark & \checkmark & & \checkmark & & & \checkmark \\
   Carreras et al. 2002 & \checkmark & \checkmark & \checkmark & \checkmark & & & & \checkmark & \checkmark \\
   Clark and Curran 2003 & \checkmark & \checkmark &  & \checkmark & & & & \checkmark & \checkmark \\
   Sonar nerd & \checkmark & \checkmark & & \checkmark & & \checkmark & \checkmark & \checkmark & \checkmark \\
   Alegria et al. 2006 & \checkmark & \checkmark &  & \checkmark &  & \checkmark & \checkmark \\
   ixa-pipe-nerc & \checkmark & & & \checkmark & \checkmark & & & \checkmark & \checkmark \\ \hline
  \end{tabular}
  \caption{Features of best previous in-domain results. Local: shallow local features including capitalization, word shape, etc.; Ling: linguistic features such as POS, lemma, chunks and semantic information from Wordnet; Global: global features; Gaz: gazetteers; WR: word representation features; Rules: manually encoded rules; Ensemble: stack of classifiers or ensemble system; Public: if the system is publicly distributed. Res: If any external resources used are publicly distributed to allow re-training.}
  \label{tab:features}
\end{table*}

Traditionally, local features have included contextual and orthographic information, affixes, character-based features, prediction history, etc. As argued by the CoNLL 2003 organizers, no feature set was deemed to be ideal for NERC \citep{tjong_kim_sang_introduction_2003}, although many approaches for English refer to \cite{zhang2003robust} as a useful general approach.

\paragraph{Linguistic Information}

Some of the CoNLL participants use linguistic information (POS, lemmas, chunks, but also specific rules or patterns) for Dutch and English \citep{carreras_named_2002,ClarkCurran:2003}, although these type of features was deemed to be most important for German, for which the use of linguistic features is pervasive \citep{benikova7germeval}. This is caused by the sparsity caused by the declension cases, the tendency to form compounds containing named entities and by the capitalization of every noun \citep{faruqui_training_2010}. For example, the best system among the 11 participants in GermEval 2014, ExB, uses morphological features and specific suffix lists aimed at capturing frequent patterns in the endings of named entities \citep{exb}.

In agglutinative languages such as Basque, which contains declension cases for named entities, linguistic features are considered to be a requirement. For example, the country name `Espainia' (Spain in Basque) can occur in several forms, \emph{Espainian}, \emph{Espainiera}, \emph{Espainiak}, \emph{Espainiarentzat}, \emph{Espainiako}, and many more.\footnote{English: in Spain, to Spain, Spain (in transitive clause), for Spain, in Spain.} Linguistic information has been used to treat this phenomenon. The only previous work for Basque developed Eihera, a rule-based NERC system formalized as finite state transducers to take into account declension classes \citep{alegria2006lessons}. The features of Eihera include word, lemma, POS, declension case, capitalized lemma, etc. These features are complemented with gazetteers extracted from the \emph{Euskaldunon Egunkaria} newspaper and semantic information from the Basque WordNet.

\paragraph{Gazetteers}

Dictionaries are widely used to inject world knowledge via gazetteer matches as features in machine learning approaches to NERC. The best performing systems carefully compile their own gazetteers from a variety of sources \citep{carreras_named_2002}. \cite{ratinov_design_2009} leverage a collection of 30 gazetteers and matches against each one are weighted as a separate feature. In this way they trust each gazetteer to a different degree. \cite{passos-kumar-mccallum:2014:W14-16} carefully compiled a large collection of English gazetteers extracted from US Census data and Wikipedia and applied them to the process of inducing word embeddings with very good results.

While it is possible to automatically extract them from various corpora or resources, they still require careful manual inspection of the target data. Thus, our approach only uses off the shelf gazetteers whenever they are publicly available. Furthermore, our method collapses every gazetteer into one dictionary. This means that we only add a feature per token, instead of a feature per token and gazetteer.

\paragraph{Global Features}

The intuition behind non-local (or global) features is to treat similarly all occurrences of the same named entity in a text. \cite{carreras_named_2002} proposed a method to produce the set of named entities for the whole sentence, where the optimal set of named entities for the sentence is the coherent set of named entities which maximizes the summation of confidences of the named entities in the set. \cite{ratinov_design_2009} developed three types of non-local features, analyzing global dependencies in a window of between 200 and 1000 tokens.

\paragraph{Word representations}

Semi-supervised approaches leveraging unlabeled text had already been applied to improve results in various NLP tasks. More specifically, it had been previously shown how to apply Brown clusters \citep{brown1992class} for Chinese Word Segmentation \citep{liang2005semi}, dependency parsing \citep{koo-carreras-collins:2008:ACLMain}, NERC \citep{suzuki-isozaki:2008:ACLMain} and POS tagging \citep{biemann2009unsupervised}.

\cite{ratinov_design_2009} used Brown clusters as features obtaining what was at the time the best published result of an English NERC system on the CoNLL 2003 testset. \cite{turian-ratinov-bengio:2010:ACL} made a rather exhaustive comparison of Brown clusters, Collobert and Weston's embeddings \citep{collobert2008unified} and HLBL embeddings \citep{mnih2007three} to improve chunking and NERC. They show that in some cases the combination of word representation features was positive but, although they used Ratinov and Roth's (2009) system as starting point, they did not manage to improve over the state of the art. Furthermore, they reported that Brown clustering features performed better than the word embeddings.

\cite{passos-kumar-mccallum:2014:W14-16} extend the Skip-gram algorithm to learn 50-dimensional lexicon infused phrase embeddings from 22 different gazetteers and the Wikipedia. The resulting embeddings are used as features by scaling them by a hyper-parameter which is a real number tuned on the development data. \cite{passos-kumar-mccallum:2014:W14-16} report best results up to date for English NERC on CoNLL 2003 test data, 90.90 F1.

The best German CoNLL 2003 system (an ensemble) was outperformed by \cite{faruqui_training_2010}. They trained the Stanford NER system \citep{finkel_incorporating_2005}, which uses a linear-chain Conditional Random Field (CRF) with a variety of features, including lemma, POS tag, etc. Crucially, they included ``distributional similarity'' features in the form of  Clark clusters \citep{clark2003combining} induced from large unlabeled corpora: the Huge German Corpus (HGC) of around 175M tokens of newspaper text and the deWac corpus \citep{baroni2009wacky} consisting of 1.71B tokens of web-crawled data. Using the clusters induced from deWac as a form of \emph{semi-supervision} improved the results over the best CoNLL 2003 system by 4 points in F1.

\paragraph{Ensemble Systems}

The best participant of the English CoNLL 2003 shared task used the results of two externally trained NERC taggers to create an ensemble system \citep{florian2003conll}. \cite{passos-kumar-mccallum:2014:W14-16} develop a stacked linear-chain CRF system: they train two CRFs with roughly the same features; the second CRF can condition on the predictions made by the first CRF. Their ``baseline'' system uses a similar local featureset as Ratinov and Roth's (2009) but complemented with gazetteers. Their baseline system combined with their phrase embeddings trained with infused lexicons allow them to report the best CoNLL 2003 result so far.

The best system of the GermEval 2014 task built an ensemble of classifiers and pattern extractors to find the most likely tag sequence \citep{exb}. They paid special attention to out of vocabulary words which are addressed by semi-supervised word representation features and an ensemble of POS taggers. Furthermore, remaining unknown candidate mentions are tackled by look-up via the Wikipedia API.

Apart from the feature types, the last two columns of Table \ref{tab:features} refer to whether the systems are publicly available and whether any external resources used for training are made available (e.g., induced word embeddings, gazetteers or corpora). This is desirable to be able to re-train the systems on different datasets. For example, we would have been interested in training the Stanford NER system with the full Ancora corpus for the evaluation presented in Table \ref{tab:classout}, but their Spanish cluster lexicon is not available. Alternatively, we would have liked to train our system with the same Ancora partition used to train Stanford NER, but that is not available either.

\section{System Description}\label{sec:system-description}

The design of \emph{ixa-pipe-nerc} aims at establishing a simple and shallow feature set, avoiding any linguistic motivated features, with the objective of removing any reliance on costly extra gold annotations (POS tags, lemmas, syntax, semantics) and/or cascading errors if automatic language processors are used. The underlying motivation is to obtain robust models to facilitate the development of NERC systems for other languages and datasets/domains while obtaining state of the art results. Our system consists of:

\begin{itemize}
\item Local, shallow features based mostly on orthographic, word shape and n-gram features plus their context.
\item Three types of simple clustering features, based on unigram matching.
\item Publicly available gazetteers, widely used in previous NERC systems \citep{tjong_kim_sang_introduction_2003,nadeau_survey_2007}.
\end{itemize}

Table \ref{tab:ekuadorko} provides an example of the features generated by our system\footnote{To avoid too much repetition, Brown, Trigram and character n-gram features have been abbreviated.}.

\begin{table*}[ht]\footnotesize
  \centering
  \begin{tabular}{llllll} \hline
   Feature & $w_{i-2}$ & $w_{i-1}$ & $w_i$ & $w_{i+1}$ & $w_{i+2}$ \\ \hline \hline
   Token & w=1994an & w=, & w=ekuadorko & w=hiriburuan & w=, \\
   Token Shape & wc=1994an,4d & wc=,,other & wc=ekuadorko,ic &  wc=hiriburuan,lc & wc=,,other \\
   Previous Pred & pd=null & pd=other & pd=null & pd=null & pd=other \\ \hline
   \multirow{2}{*}{Brown Token} & bt=0111 & & bt=0010 & bt=0101 \\
   & bt=011111 & & bt=001001 & bt=010110 \\ \hline
   \multirow{2}{*}{Brown Token,Class} & c,bt=4d,0111 & & c,bt=ic,0010 & c,bt=lc,0101 \\
   & c,bt=4d,011111 & & c,bt=ic,001001 & c,bt=lc,010111 \\ \hline
   Clark-a & ca=158 & ca=O & ca=175 & ca=184 & ca=O \\
   Clark-b & cb=149 & cb=O & cb=176 & cb=104 & cb=O \\ \hline
   Word2vec-a & w2va=55 & w2va=O & w2va=14 & w2va=14 & w2va=O \\
   Word2vec-b & w2vb=524 & w2vb=O & w2vb=464 & w2vb=139 & w2vb=O \\ \hline
   Prefix($w_i$) & \multicolumn{5}{l}{pre=Eku; pre=Ekua} \\
   Suffix($w_i$) & \multicolumn{5}{l}{suf=o; suf=ko; suf=rko; suf=orko} \\
   Bigram($w_i$) & \multicolumn{5}{l}{pw,w=,,Ekuadorko; pwc,wc=other,ic; w,nw=Ekuadorko,hiriburuan; wc,nc=ic,lc} \\
   Trigram($w_i$) & \multicolumn{5}{l}{ppw,pw,w=1994an,,,Ekuadorko; ppwc,pwc,wc=4d,other,ic; $\ldots$}\\
   char n-grams($w_i$) & \multicolumn{5}{l}{ng=adorko; ng=rko; ng=dorko; ng=ko; ng=orko $\ldots$ } \\ \hline
  \end{tabular}
  \caption{Features generated for the Basque sentence ``Morras munduko txapeldun izan zen juniorretan 1994an, Ekuadorko hiriburuan, Quiton''. English: Morras was junior world champion in 1994, in the capital of Ecuador, Quito. Current token is `Ekuadorko'. }
  \label{tab:ekuadorko}
\end{table*}

\subsection{Local Features}\label{sec:local-features}

The local features constitute our baseline system on top of which the clustering features are added. We implement the following feature set, partially inspired by previous work \citep{zhang2003robust}:

\begin{itemize}
\item Token: Current lowercase token (w), namely, \emph{ekuadorko} in Table \ref{tab:ekuadorko}.
\item Token Shape: Current lowercase token (w) \emph{plus} current token shape (wc), where token shape consist of: (i) The token is either lowercase or a 2 digit word or a 4 digit word; (ii) If the token contains digits, then whether it also contains letters, or slashes, or hyphens, or commas, or periods or is numeric; (iii) The token is all uppercase letters \emph{or} is an acronym \emph{or} is a one letter uppercase word \emph{or} starts with capital letter.
Thus, in Table \ref{tab:ekuadorko} \emph{1994an} is a 4 digit word (4d), \emph{Ekuadorko} has an initial capital shape (ic) and \emph{hiriburuan} is lowercase (lc).
\item Previous prediction: the previous outcome (pd) for the current token. The previous predictions in our example are \emph{null} because these words have not been seen previously, except for the comma.
\item Sentence: Whether the token is the beginning of the sentence. None of the tokens in our example is at the beginning of the sentence, so this feature is not active in Table \ref{tab:ekuadorko}.
\item Prefix: Two prefixes consisting of the first three and four characters of the current token: \emph{Eku} and \emph{Ekua}.
\item Suffix: The four suffixes of length one to four from the last four characters of the current token.
\item Bigram: Bigrams including the current token and the token shape.
\item Trigram: Trigrams including the current token and the token shape.
\item Character n-gram: All lowercase character bigrams, trigrams, fourgrams and fivegrams from the current token (ng).
\end{itemize}

Token, token shape and previous prediction features are placed in a 5 token window, namely, for these these three features we also consider the previous and the next two words, as shown in Table \ref{tab:ekuadorko}.

\subsection{Gazetteers}\label{sec:gazetteers}

We add gazetteers to our system only if they are readily available to use, but our approach does not fundamentally depend upon them. We perform a look-up in a gazetteer to check if a named entity occurs in the sentence. The result of the look-up is represented with the same encoding chosen for the training process, namely, the BIO or BILOU scheme\footnote{The BIO scheme suggests to learn models that identify the Beginning, the Inside and the Outside of sequences. The BILOU scheme proposes to learn models Beginning, the
Inside and the Last tokens of multi-token chunks as well as Unit-length chunks.}. Thus, for the current token we add the following features:

\begin{enumerate}
\item The current named entity class in the encoding schema. Thus, in the BILOU encoding we would have ``unit'', ``beginning'', ``last'', ``inside'', or if not match is found, ``outside'', combined with the specific named entity type (LOC, ORG, PER, MISC, etc.).
\item The current named entity class as above \emph{and} the current token.
\end{enumerate}

\subsection{Clustering Features}\label{sec:clustering-features}

The general idea is that by using some type of semantic similarity or word cluster induced over large unlabeled corpora it is possible to improve the predictions for unseen words in the test set. This type of semi-supervised learning may be aimed at improving performance over a fixed amount of training data or, given a fixed target performance level, to establish how much supervised data is actually required to reach such performance \citep{koo-carreras-collins:2008:ACLMain}.

So far the most successful approaches have only used one type of word representation \citep{passos-kumar-mccallum:2014:W14-16,faruqui_training_2010,ratinov_design_2009}. However, our simple baseline combined with one type of word representation features are not able to compete with previous, more complex, systems. Thus, instead of encoding more elaborate features, we have devised a simple method to \emph{combine} and \emph{stack} various types of clustering features induced over different data sources or corpora. In principle, our method can be used with any type of word representations. However, for comparison purposes, we decided to use word representations previously used in successful NERC approaches: Brown clusters \citep{ratinov_design_2009,turian-ratinov-bengio:2010:ACL}, Word2vec clusters \citep{passos-kumar-mccallum:2014:W14-16} and Clark clusters \citep{finkel_incorporating_2005,faruqui_training_2010}. As can be observed in Table \ref{tab:ekuadorko}, our clustering features are placed in a 5 token window.

\subsubsection{Brown Features}\label{sec:brown-clusters}

The Brown clustering algorithm \citep{brown1992class} is a hierarchical algorithm which clusters words to maximize the mutual information of bigrams. Thus, it is a class-based bigram model in which:

\begin{itemize}
\item The probability of a document corresponds to the product of the probabilities of its bigrams,
\item the probability of each bigram is calculated by multiplying the probability of a bigram model over latent classes by the probability of each class generating the actual word types in the bigram, and
\item each word type has non-zero probability only on a single class.
\end{itemize}

\begin{figure}[ht]
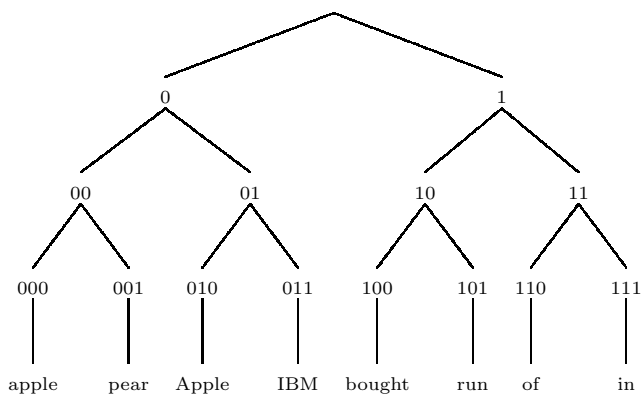

  \centering
\scriptsize{\synttree[ [0 [00 [000 [apple]] [001 [pear]]] [01 [010 [Apple]] [011 [IBM]]] ] [1 [10 [100 [bought] ] [101 [run] ]] [11 [110 [of]] [111 [in]] ] ] ] }
  \caption{A Brown clustering hierarchy.}
  \label{fig:brown}
\end{figure}

The Brown algorithm takes a vocabulary of words to be clustered and a corpus of text containing these words. It starts by assigning each word in the vocabulary to its own separate cluster, then iteratively merges the pair of clusters which leads to the smallest decrease in the likelihood of the text corpus. This produces a hierarchical clustering of the words, which is usually represented as a binary tree, as shown in Figure \ref{fig:brown}. In this tree every word is uniquely identified by its path from the root, and the path can be represented by a bit string. It is also possible to choose different levels of word abstraction by choosing different depths along the path from the root to the word. Therefore, by using paths of various lengths, we obtain clustering features of different granularities \citep{miller2004name}.

We use paths of length 4, 6, 10 and 20 as features \citep{ratinov_design_2009}. However, we introduce several novelties in the design of our Brown clustering features:

\begin{enumerate}
\item For each feature which is token-based, we add a feature containing the paths computed for the current token. Thus, taking into account our baseline system, we will add the following Brown clustering features:
  \begin{enumerate}
  \item Brown Token: existing paths of length 4, 6, 10 and 20 for the current token.
  \item Brown Token Shape: existing paths of length 4, 6, 10, 20 for the current token \emph{and} current token shape.
  \item Brown Bigram: existing paths of length 4, 6, 10, 20 for bigrams including the current token.
  \end{enumerate}
\item Brown clustering features benefit from two additional features:
  \begin{enumerate}
  \item Previous prediction plus token: the previous prediction (pd) for the current token \emph{and} the current token.
  \item Previous two predictions: the previous prediction for the current \emph{and} the previous token.
  \end{enumerate}
\end{enumerate}

For space reasons, Table \ref{tab:ekuadorko} only shows the Brown Token (bt) and Brown Token Shape (c) features for paths of length 4 and 6. We use the publicly available tool\footnote{\url{https://github.com/percyliang/brown-cluster}} implemented by \cite{liang2005semi} with default settings. The input consists of a corpus tokenized and segmented one sentence per line, without punctuation. Furthermore, we follow previous work and remove all sentences which consist of less than 90\% lowercase characters \citep{liang2005semi,turian-ratinov-bengio:2010:ACL} before inducing the Brown clusters.

\subsubsection{Clark Features}\label{sec:clark-clusters}

\cite{clark2003combining} presents a number of unsupervised algorithms, based on distributional and morphological information, for clustering words into classes from unlabeled text. The focus is on clustering infrequent words on a small numbers of clusters from comparatively small amounts of data. In particular, \cite{clark2003combining} presents an algorithm combining distributional information with morphological information of words ``by composing the Ney-Essen clustering model with a model for the morphology within a Bayesian framework''. The objective is to bias the distributional information to put words that are morphologically similar in the same cluster. We use the code released by \cite{clark2003combining} off the shelf\footnote{\url{https://github.com/ninjin/clark_pos_induction}} to induce Clark clusters using the Ney-Essen with morphological information method. The input of the algorithm is a sequence of lowercase tokens without punctuation, one token per line with sentence breaks.

Our Clark clustering features are very simple: we perform a look-up of the current token in the clustering lexicon. If a match is found, we add as a feature the clustering class, or the lack of match if the token is not found (see Clark-a and Clark-b in Table \ref{tab:ekuadorko}).

\subsubsection{Word2vec Features}\label{sec:Word2vec-clusters}

Another family of language models that produces word representations are the neural language models. These approaches produce representation of words as continuous vectors \citep{collobert2008unified,mnih2007three}, also called \emph{word embeddings}. Nowadays, perhaps the most popular among them is the Skip-gram algorithm \citep{mikolov2013distributed}. The Skip-gram algorithm uses shallow log-linear models to compute vector representation of words which are more efficient than previous word representations induced on neural language models. Their objective is to produce word embeddings by computing the probability of each n-gram as the product of the conditional probabilities of each context word in the n-gram conditioned on its central word \citep{mikolov2013distributed}.

Instead of using continuous vectors as real numbers, we induce clusters or word classes from the word vectors by applying K-means clustering. In this way we can use the cluster classes as simple binary features by injecting unigram match features. We use the Word2vec tool\footnote{\url{https://code.google.com/p/Word2vec/}} released by \cite{mikolov2013distributed} with a 5 window context to train 50-dimensional word embeddings and to obtain the word clusters on top of them. The input of the algorithm is a corpus tokenized, lowercased, with punctuation removed and in one line. The Word2vec features are implemented exactly like the Clark features.

\subsubsection{Stacking and Combining Clustering Features}\label{sec:comb-clust-feat}

We successfully combine clustering features from different word representations. Furthermore, we also \emph{stack} or \emph{accumulate} features of the same type of word representation induced from different data sources, trusting each clustering lexicon to a different degree, as shown by the five encoded clustering features in Table \ref{tab:ekuadorko}: two Clark and Word2vec features from different source data and one Brown feature. When using word representations as semi-supervised features for a task like NERC, two principal factors need to be taken into account: (i) the source data or corpus used to induce the word representations and (ii) the actual word representation used to encode our features which in turn modify the weight of our model's parameters in the training process.

For the clustering features to be effective the induced clusters need to contain as many words appearing in the training, development and test sets as possible. This can be achieved by using corpora closely related to the text genre or domain of the data sets or by using very large unlabeled corpora which, although not closely domain-related, be large enough to include many relevant words. For example, with respect to the CoNLL 2003 English dataset an example of the former would be the Reuters corpus while the Wikipedia would be an example of the latter.

The word representations obtained by different algorithms would capture different distributional properties of words in a given corpus or data source. Therefore, each type of clustering would allow us to capture different types of occurring named entity types. In other words, \emph{combining} and \emph{stacking} different types of clustering features induced over a variety of data sources should help to capture more similarities between different words in the training and test sets, increasing the contribution to the weights of the model parameters in the training process.

\section{Experimental Results}\label{sec:results}

In this Section we report on the experiments performed with the \emph{ixa-pipe-nerc} system as described in the previous section. The experiments are performed in 5 languages: Basque, Dutch, English, German and Spanish. For comparison purposes, in-domain results are presented in Section \ref{sec:domain-evaluations} using the most common NERC datasets for each language as summarized in Table \ref{tab:datasets}. Section \ref{sec:reduc-train-data} analyzes the performance when reducing training data and Section \ref{sec:out-doma-eval} presents eight out-of-domain evaluations for three languages: Dutch, English and Spanish.

\begin{table*}[ht]\footnotesize
  \centering
  \begin{tabular}{clrrrr} \hline
   & \multicolumn{2}{c}{million words in corpus} & \multicolumn{3}{r}{million words for training}\\ \hline \hline
   & & & Brown & Clark & Word2vec \\ \hline
   \multirow{3}{*}{en} & Reuters RCV1 & 63 & 35 & 63 & 63 \\
    & Wikipedia (20141208) & 1700 & 790 & 790 & 1700 \\
    & Gigaword 5th ed. & 4000 & - & - & 4000 \\ \hline
   \multirow{2}{*}{de} & Wikipedia (20140725) & 650 & 190 & 190 & 650 \\
    & deWac \citep{baroni2009wacky} & 1100 & 500 & 500 & 1100 \\ \hline
   \multirow{3}{*}{es} & Wikipedia (20140810) & 428 & 246 & 246 & 428 \\
    & elperiodico (1998-2002) & 60 & 35 & 60 & 60 \\
    & Gigaword 3rd ed. & 1150 & 330 (afp) & 330 (afp) & 1150 \\ \hline
   nl & Wikipedia (20140804) & 235 & 128 & 128 & 235 \\ \hline
   \multirow{3}{*}{eu} & Wikipedia (20141208) & 60 & 12 & 60 & 60 \\
    & Egunkaria (1999-2003) & 38 & 28 & 38 & 38 \\
    & Berria (2003-2014) & 90 & 78 & 90 & 90 \\ \hline
  \end{tabular}
  \caption{Unlabeled corpora used to induced clusters. For each corpus and cluster type the number of words (in millions) is specified. Average training times: depending on the number of words, Brown clusters training time required between 5h and 48h. Word2vec required 1-4 hours whereas Clark clusters training lasted between 5 hours and 10 days.}
  \label{tab:unlabeledcorpora}
\end{table*}

The results for Dutch, English and Spanish do not include trigrams and character n-grams in the local featureset described in Section \ref{sec:local-features}, except for the models in each in-domain evaluation which are marked with ``charngram 1:6''.

We also experiment with dictionary features but, in contrast to previous approaches such as \cite{passos-kumar-mccallum:2014:W14-16}, we only use currently available gazetteers off-the-shelf. For every model marked with ``dict'' we use the thirty English Illinois NER gazetteers \citep{ratinov_design_2009}, irrespective of the target language. Additionally, the English models use six gazetteers about the Global Automotive Industry provided by LexisNexis to the Newsreader project\footnote{\url{http://www.newsreader-project.eu}}, whereas the German models include, in addition to the Illinois gazetteers, the German dictionaries distributed in the CoNLL 2003 shared task. The gazetteers are collapsed into one large dictionary and deployed as described in Section \ref{sec:gazetteers}.

Finally, the clustering features are obtained by processing the following clusters from publicly available corpora: (i) 1000 Brown clusters; (ii) Clark and Word2vec clusters in the 100-600 range. To choose the best combination of clustering features we test the available permutations of Clark and Word2vec clusters with and without the Brown clusters on the development data. Table \ref{tab:unlabeledcorpora} provides details of every corpus used to induce the clusters. For example, the first row reads: ``Reuters RCV1 was used; the original 63 million words were reduced to 35 million after pre-processing for inducing Brown clusters. Clark and Word2vec clusters were trained on the whole corpus''. The pre-processing and tokenization is performed with the IXA pipes tools \citep{AGERRI14.775.L14-1605}.

Every evaluation is carried out using the CoNLL NER evaluation script\footnote{\url{http://www.cnts.ua.ac.be/conll2002/ner/bin/conlleval.txt}}. The results are obtained with the BILOU encoding for every experimental setting except for German CoNLL 2003.

\subsection{In-domain evaluation}\label{sec:domain-evaluations}

In this section the results are presented by language. In two cases, Dutch and German, we use two different datasets, making it a total of seven in-domain evaluations.

\subsubsection{English}\label{sec:english}

We tested our system in the highly competitive CoNLL 2003 dataset. Table \ref{tab:englishconll03} shows that three of our models outperform previous best results reported for English in the CoNLL 2003 dataset \citep{passos-kumar-mccallum:2014:W14-16}. Note that the best F1 score (91.36) is obtained by adding trigrams and character n-gram features to the best model (91.18). The results also show that these models improve the baseline provided by the local features by around 7 points in F1 score. The most significant gain is in terms of recall, almost 9 points better than the baseline.

\begin{table*}[ht]\footnotesize
  \centering
  \begin{tabular}{lcccccc} \hline
   & \multicolumn{3}{c}{Development} & \multicolumn{3}{c}{Test} \\ \hline \hline
    Features & P & R & F1 & P & R & F1 \\ \hline
    Local (L) & 93.02 & 87.75 & 90.31 & 87.27 & 81.32 & 84.19 \\
    L + Brown reuters (BR) & 92.83 & 89.33 & 91.05 & 90.28 & 86.79 & 88.50 \\
    L + Clark wiki 600 (CW600) & 93.98 & 90.58 & 92.24 & 90.85 & 87.16 & 88.97 \\
    L + Word2vec giga 200 (W2VG200) & 93.16 & 89.90 & 91.45 & 89.64 & 85.06 & 87.29 \\
    L + Word2vec wiki 400 (W2VW400) & 93.22 & 90.02 & 91.59 & 88.98 & 85.09 & 86.99 \\
    L + BR + CW600 + W2VW400 (light) & 94.16 & 91.96 & 93.04 & 91.20 & 89.36 & 90.27 \\
    light + CR600 + W2VG200 (comp) & 94.32 & 92.22 & 93.26 & 91.75 & 89.64 & 90.69 \\
    comp + BW (best cluster) & 94.21 & 92.23 & 93.26 & 91.67 & 89.98 & 90.82 \\ \hline
    comp + dict & \textbf{94.60} & 92.78 & 93.68 & 91.86 & 90.53 & \textbf{91.19} \\
    BR+CR600-CW600+W2VG200+dict & 94.58 & 92.53 & 93.54 & \textbf{92.20} & 90.19 & \textbf{91.18} \\
    charngram 1:6 + en-91-18 & 94.56 & 92.81 & 93.68 & 92.16 & 90.56 & \textbf{91.36} \\ \hline
    Stanford NER (distsim-conll03) & 93.64 & 92.27 & 92.95 & 89.37 & 87.95 & 88.65 \\
    Illinois NER & - & - & 93.50 & n/a & n/a & 90.57 \\
    Turian et al. (2010) & 94.11 & \textbf{93.81} & 93.95 & 90.10 & \textbf{90.61} & 90.36 \\
    Passos et al. (2014) & - & - & \textbf{94.46} & - & - & 90.90 \\ \hline
  \end{tabular}
  \caption{CoNLL 2003 English results.}
  \label{tab:englishconll03}
\end{table*}

We also report very competitive results, only marginally lower than \cite{passos-kumar-mccallum:2014:W14-16}, based on the \emph{stacking} and \emph{combination} of clustering features  as described in Section \ref{sec:comb-clust-feat}. Thus, both \emph{best cluster} and \emph{comp} models, based on local plus clustering features only, outperform very competitive and more complex systems such as those of \cite{ratinov_design_2009} and \cite{turian-ratinov-bengio:2010:ACL}, and obtain only marginally lower results than \cite{passos-kumar-mccallum:2014:W14-16}. The \emph{stacking} and \emph{combining} effect manifests itself very clearly when we compare the single clustering feature models (BR, CW600, W2VG200 and W2VW400) with the \emph{light}, \emph{comp} and \emph{best cluster} models which improve the overall F1 score by 1.30, 1.72 and 1.85 respectively over the best single clustering model (CW600).

It is worth mentioning that our models do not score best in the development data. As the development data is closer in style and genre to the training data \citep{ratinov_design_2009}, this may suggest that our system generalizes better on test data that is not close to the training data; indeed, the results reported in Section \ref{sec:out-doma-eval} seem to confirm this hypothesis.

We also compared our results with respect to the best two publicly available English NER systems trained on the same data. We downloaded the Stanford NER system distributed in the 2015-01-30 package. We evaluated their CoNLL model and, while the result is substantially better than their reference paper \citep{finkel_incorporating_2005}, our clustering models obtain better results. The Illinois NER tagger is used by \cite{ratinov_design_2009} and \cite{turian-ratinov-bengio:2010:ACL}, both of which are outperformed by our system.

\subsubsection{German}\label{sec:german}

\begin{table*}[ht]\footnotesize
  \centering
  \begin{tabular}{lcccccc} \hline
   & \multicolumn{3}{c}{Outer NEs} & \multicolumn{3}{c}{Inner NEs} \\ \hline \hline
    Features & P & R & F1 & P & R & F1 \\ \hline
    Local (L) & 74.64 & 67.35 & 70.81 & 61.30 & 34.76 & 44.36 \\
    L + Brown wiki (BW) & 79.71 & 73.20 & 76.31 & 63.82 & 37.67 & 47.37 \\
    L + Clark wiki 200 (CW200) & 78.86 & 72.00 & 75.27 & \textbf{65.49} & 36.12 & 46.56 \\
    L + Word2vec wiki 200 (W2VW200) & 78.00 & 71.75 & 74.75 & 62.62 & 38.06 & 47.34 \\
    L + Word2vec deWac 100 (W2VWac100) & 71.36 & 71.63 & 74.38 & 62.84 & 36.12 & 45.87 \\
    BW+CW200+W2V(W200+Wac100) (de-cluster) & 81.00 & 75.14 & 77.96 & 62.80 & 40.00 & \textbf{48.87} \\ \hline
    de-cluster + dict & \textbf{81.52} & 75.54 & 78.42 & 60.28 & 41.55 & \textbf{49.20} \\ \hline
    ExB \citep{exb} & 80.67 & \textbf{77.55} & \textbf{79.08} & 45.20 & 41.17 & 43.09 \\
    UKP \citep{ukp} & 79.90 & 74.13 & 76.91 & 58.74 & \textbf{41.75} & 48.81 \\ \hline
    de-4-class-cluster & 82.49 & 75.96 & 79.09 & - & - & - \\
    de-4-class-cluster + dict & 82.70 & 76.44 & 79.45 & - & - & - \\
    GermaNER (4 classes) & 82.72 & 71.19 & 76.52 & - & - & - \\ \hline
  \end{tabular}
  \caption{GermEval 2014 M3 metric results and comparison to GermaNER system on the outer spans.}
  \label{tab:germeval2014}
\end{table*}

We tested our system in the GermEval 2014 dataset. Table \ref{tab:germeval2014} compares our results with the best two systems (ExB and UKP) by means of the M3 metric, which separately analyzes the performance in terms of the outer and inner named entity spans. Table \ref{tab:germeval2014} makes explicit the significant improvements achieved by the clustering features on top of the baseline system, particularly in terms of recall (almost 11 points in the outer level). The official results of our best configuration (de-cluster-dict) are reported in Table \ref{tab:germevalofficial} showing that our system marginally improves the best systems' results on that task (ExB and UKP).

We also compare our system, in the last three rows, with the publicly available GermaNER \citep{germaner}, which reports results for the 4 main outer level entity types (person, location, organization and other). For this experiment we trained the \emph{de-cluster} and \emph{de-cluster + dict} models on the four main classes, improving GermaNER's results by almost 3 F1 points. The GermaNER method of evaluation is interesting because allows researchers to directly compare their systems with a publicly available system trained on GermEval data.

\begin{table}[ht]\footnotesize
  \centering
  \begin{tabular}{lccc} \hline
    Features & Precision & Recall & F1 \\ \hline \hline
    de-cluster + dict & \textbf{80.28} & 72.93 & \textbf{76.43} \\
    ExB & 78.07 & \textbf{74.75} & 76.38 \\
    UKP & 79.54 & 71.10 & 75.09 \\ \hline
  \end{tabular}
  \caption{GermEval 2014 Official results.}
  \label{tab:germevalofficial}
\end{table}

\begin{table*}[ht]\footnotesize
  \centering
  \begin{tabular}{lcccccc} \hline
   & \multicolumn{3}{c}{Development} & \multicolumn{3}{c}{Test} \\ \hline \hline
    Features & P & R & F1 & P & R & F1 \\ \hline
    Local (L) & 79.38 & 55.27 & 65.16 & 81.70 & 59.92 & 69.14 \\
    L + Brown deWac (BWac) & 82.12 & 66.73 & 73.63 & 82.26 & 66.53 & 73.57 \\
    L + Clark deWac 500 (CWac500) & 82.40 & 65.69 & 73.11 & 82.76 & 67.33 & 74.25 \\
    L + Word2vec deWac 100 (W2VWac100) & 81.50 & 63.52 & 71.40 & 81.96 & 66.46 & 73.41 \\
    CWac500 + W2VWac100 (de-cluster) & 83.35 & 68.38 & 75.13 & 83.43 & 68.96 & \textbf{75.51} \\ \hline
    de-cluster + dict & 85.20 & \textbf{70.54} & 77.18 & 83.72 & \textbf{70.30} & \textbf{76.42} \\ \hline
    Florian et al. (2003) & 84.60 & 61.93 & 71.51 & 80.19 & 63.71 & 72.41 \\
    Faruqui and Pad\'o (2010) & \textbf{86.00} & 70.00 & \textbf{77.20} & \textbf{86.40} & 68.50 & 76.40 \\ \hline
  \end{tabular}
  \caption{CoNLL 2003 German results.}
  \label{tab:germanconll03}
\end{table*}

Table \ref{tab:germanconll03} compares our German CoNLL 2003 results with the best previous work trained on public data. Our best CoNLL 2003 model obtains results similar to the state of the art performance with respect to the best system published up to date \citep{faruqui_training_2010} using public data. \noindent \cite{faruqui_training_2010} also report 78.20 F1 with a model trained with Clark clusters induced using the Huge German Corpus (HGC). Unfortunately, the corpus or the induced clusters were not available.

\subsubsection{Spanish}\label{sec:spanish}

The best system up to date on the CoNLL 2002 dataset, originally published by \cite{carreras_named_2002}, is distributed as part of the Freeling library \citep{freeling3_padro12}. Table \ref{tab:spanishconll02} lists four models that improve over their reported results, almost by 3 points in F1 measure in the case of the \emph{es-cluster} model (with our without trigram and character n-gram features).

\begin{table*}[ht]\footnotesize
  \centering
  \begin{tabular}{lcccccc} \hline
   & \multicolumn{3}{c}{Development} & \multicolumn{3}{c}{Test} \\ \hline \hline
    \textbf{Features} & P & R & F1 & P & R & F1 \\ \hline
    Local (L) & 77.78 & 75.30 & 76.52 & 79.49 & 79.54 & 79.52\\
    L + Brown periodico (BP) & 80.35 & 78.91 & 79.62 & 82.44 & 82.46 & 82.45 \\
    L + Clark giga 400 (CG400) & 81.71 & 79.27 & 80.48 & 81.75 & 81.82 & 81.78 \\
    L + Clark wiki 400 (CW400) & 80.84 & 78.75 & 79.78 & 81.07 & 81.00 & 81.03 \\
    L + Word2vec giga 400 (W2VG400) & 80.04 & 77.96 & 78.99 & 81.56 & 81.76 & 81.67 \\
    BP+C(W400+G400)+W2VG400 (es-cluster) & 81.87 & 80.61 & 81.23 & 84.18 & \textbf{84.15} & \textbf{84.16} \\
    charngram 1:6 + es-cluster & 81.95 & 80.24 & 81.09 & \textbf{84.27} & 84.01 & 84.14 \\ \hline
    Carreras et al. (2002) & 79.15 & 77.80 & 78.47 & 81.38 & 81.40 & 81.39 \\ \hline
  \end{tabular}
  \caption{CoNLL 2002 Spanish results.}
  \label{tab:spanishconll02}
\end{table*}

\subsubsection{Dutch}\label{sec:dutch}

Despite using clusters from one data source only (see Table \ref{tab:unlabeledcorpora}), results in Table \ref{tab:dutchconll02} show that our \emph{nl-cluster} model outperforms the best result published on CoNLL 2002 \citep{ClarkCurran:2003} by 3.83 points in F1 score. Adding the English Illinois NER gazetteers \citep{ratinov_design_2009} and trigram and character n-gram features increases the score to 85.04 F1, 5.41 points better than previous published work on this dataset.

\begin{table*}[ht]\footnotesize
  \centering
  \begin{tabular}{lcccccc} \hline
   & \multicolumn{3}{c}{Development} & \multicolumn{3}{c}{Test} \\ \hline \hline
    Features & P & R & F1 & P & R & F1 \\ \hline
    Local (L) & 75.66 & 70.95 & 73.23 & 78.97 & 74.80 & 76.83\\
    L + Brown wiki (BW) & 80.50 & 76.99 & 78.70 & 83.12 & 79.56 & 81.32 \\
    L + Clark wiki 400 (CW400) & 79.44 & 75.92 & 77.64 & 83.18 & 80.21 & 81.67 \\
    L + Word2vec wiki 100 (W2VW100) & 79.28 & 75.76 & 77.48 & 83.11 & 80.18 & 81.62 \\
    BW+CW400+W2VW100 (nl-cluster) & 82.16 & 79.20 & 80.65 & 84.44 & 82.49 & \textbf{83.46} \\
    nl-cluster + dict & 83.63 & 80.66 & 82.12 & 85.92 & 83.00 & \textbf{84.43} \\
    charngram 1:6 + nl-cluster & 83.27 & 80.47 & 81.84 & 85.52 & 82.42 & 83.94 \\
    charngram 1:6 + nl-cluster-dict & 84.65 & 81.57 & 83.08 & \textbf{86.57} & \textbf{83.56} & \textbf{85.04} \\ \hline
    Carreras et al. (2002) & 76.52 & 74.82 & 75.66 & 77.83 & 76.29 & 77.05 \\
    Curran and Clark (2003) & - & - & - & 79.91 & 79.35 & 79.63 \\ \hline
  \end{tabular}
  \caption{CoNLL 2002 Dutch results.}
  \label{tab:dutchconll02}
\end{table*}

We also compared our system with the more recently developed SONAR-1 corpus and the companion NERD system distributed inside its release \citep{desmet2014fine}. They report 84.91 F1 for the six main named entity types via 10-fold cross validation. For this comparison we chose the \emph{local}, \emph{nl-cluster} and \emph{nl-cluster-dict} configurations from Table \ref{tab:dutchconll02} and run them on SONAR-1 using the same settings. The results reported in Table \ref{tab:sonar-1} shows our system's improvement over previous results on this dataset.

\begin{table}[ht]\footnotesize
  \centering
  \begin{tabular}{lccc} \hline
    Features & Precision & Recall & F1 \\ \hline \hline
    Local (L) & 86.66 & 85.57 & 86.11 \\
    nl-cluster & 87.89 & 87.56 & 87.72 \\
    nl-cluster + dict & \textbf{88.08} & \textbf{87.91} & \textbf{88.00} \\ \hline
    Sonar-nerd & - & - & 84.91 \\ \hline
  \end{tabular}
  \caption{SONAR-1 10-fold cross validation results.}
  \label{tab:sonar-1}
\end{table}

\subsubsection{Basque}\label{sec:basque}

Table \ref{tab:basqueresults} reports on the experiments using the \emph{Egunkaria} NER dataset provided by \cite{alegria2006lessons}. Due to the sparsity of the MISC class mentioned in Section \ref{sec:datasets}, we decided to train our models on three classes only (location, organization and person). Thus, the results are obtained training our models in the customary manner and evaluating on 3 classes. However, for direct comparison with previous work \citep{alegria2006lessons}, we also evaluate our best \emph{eu-cluster} model (trained on 3 classes) on 4 classes.

\begin{table}[ht]\footnotesize
  \centering
  \begin{tabular}{lccc} \hline
    Features & P & R & F1 \\ \hline \hline
    Local & 70.52 & 60.27 & 65.00\\
    L + Brown egunkaria (BE) & 74.54 & 67.59 & 70.90 \\
    L + Clark egunkaria 200 (CE200) & 76.76 & 68.92 & 72.63 \\
    L + Clark wiki 200 (CW200) & 75.57 & 65.60 & 70.23 \\
    L + Word2vec egunkaria 300 (W2VE300) & 74.04 & 62.71 & 67.91 \\
    L + Word2vec berria 600 (W2WB600) & 74.11 & 64.82 & 69.15 \\
    BE+C(EW)200+ W2V(E300+B600) (eu-cluster) & 80.66 & \textbf{73.14} & \textbf{76.72} \\
    eu-cluster (4 classes) & \textbf{80.66} & 70.78 & \textbf{75.40} \\ \hline
    Alegria et al. (2006) & 72.50 & 70.24 & 71.35 \\ \hline
  \end{tabular}
  \caption{Basque Egunkaria results.}
  \label{tab:basqueresults}
\end{table}

The results show that our \emph{eu-cluster} model clearly improves upon previous work by 4 points in F1 measure (75.40 vs 71.35). These results are particularly interesting as it had been so far assumed that complex linguistic features and language-specific rules were required to perform well for agglutinative languages such as Basque \citep{alegria2006lessons}. Finally, it is worth noting that the \emph{eu-cluster} model increases the overall F1 score by 11.72 over the baseline, of which 10 points are gained in precision and 13 in terms of recall.

\subsection{Reducing training data}\label{sec:reduc-train-data}

So far, we have seen how, given a fixed amount of supervised training data, leveraging unlabeled data using multiple cluster sources helped to obtain state of the art results in seven different in-domain settings for five languages. In this section we will investigate to what extent our system allows to reduce the dependency on supervised training data.

We first use the English CoNLL 2003 dataset for this experiment. The training set consists of around 204K words and we use various smaller versions of it to test the performance of our \emph{best cluster} model reported in Table \ref{tab:englishconll03}. Table \ref{tab:eng-learn-curv} displays the F1 results of the baseline system consisting of local features and the \emph{best cluster} model. The $\Delta$ column refers to the gains of our \emph{best cluster} model with respect to the baseline model for every portion of the training set.

\begin{table}[ht]\footnotesize
  \centering
  \begin{tabular}{cccccc}\hline
\#Train Words & Local & Clusters & $\Delta$ & en-91-18 & Illinois NER \\ \hline \hline
  1/16 & 68.80 & 83.11 & 14.31 & 83.27 & 82.38 \\
  1/8 & 67.21  & 84.41 & 17.20 & 85.38 & 83.86 \\
  1/4 & 77.75 & 88.48 & 10.73 & 88.60 & 87.55 \\
  1/2 & 82.67 & 90.01 & 7.34 & 90.06 & 88.40 \\
  Full & 84.19 & 90.68 & 6.49 & 91.18 & 90.57 \\ \hline
  \end{tabular}
  \caption{CoNLL 2003 English results reducing training data.}
  \label{tab:eng-learn-curv}
\end{table}

While we have already commented the substantial gains obtained simply by adding our clustering features, it is also interesting to note that the gains are much substantial when less supervised training data is available. Furthermore, it is striking that training our clustering features using only one eight of the training data (30K words) allows to obtain similar performance to the baseline system trained on the full training set. Equally interesting is the fact that cutting by half the training data only marginally harms the overall performance. Finally, training on just a quarter of the training set (60K) results in a very competitive model when compared with other publicly available NER systems for English trained on the full training set: it roughly matches Stanford NER's performance, it outperforms models using \emph{external knowledge} or \emph{non-local features} reported by \cite{ratinov_design_2009}, and also several models reported by \cite{turian-ratinov-bengio:2010:ACL}, which use one type of word representations on top of the baseline system.

We have also re-trained the Illinois NER system \citep{ratinov_design_2009} and our best CoNLL 2003 model (\emph{en-91-18}) for comparison. First, we can observe that for every portion of the training set, both our \emph{best cluster} and \emph{en-91-18} model outperform the Illinois NER system. The \emph{best cluster} results are noteworthy because, as opposed to Illinois NER, it does not use gazetteers or global features for extra performance.

\begin{table}[ht]\footnotesize
  \centering
  \begin{tabular}{ccccccccccccc}\hline
   & \multicolumn{3}{c}{Basque} & \multicolumn{3}{c}{Dutch} & \multicolumn{3}{c}{Spanish} & \multicolumn{3}{c}{German} \\ \hline \hline
\#Train & L & C & $\Delta$ & L & C & $\Delta$ & L & C & $\Delta$ & L & C & $\Delta$ \\ \hline
1/16 & - & - & - &  59.46 & 67.63 & 8.17 & 65.07 & 71.83 & 6.76 & 51.31 & 65.97 & 14.66 \\
  1/8 & 46.27 & 62.23 & 16.96 & 63.16 & 72.41 & 9.25 & 69.24 & 74.00 & 4.76 & 56.80 & 67.84 & 11.04 \\
  1/4 & 52.15 & 65.10 & 12.95 & 71.38 & 79.25 & 7.87 & 73.79 & 78.25 & 4.46 & 61.96 & 72.20 & 10.24 \\
  1/2 & 62.13 & 71.98 & 9.85 & 74.00 & 82.46 & 8.46 & 76.87 & 80.29 & 3.42 & 66.43 & 75.45 & 9.02 \\
  Full & 65.00 & 76.72 & 11.72 & 76.83 & 83.46 & 6.63 & 79.52 & 84.16 & 4.64 & 70.81 & 77.96 & 7.15 \\ \hline
  \end{tabular}
  \caption{Multilingual results reducing training data. Datasets employed: Basque (egunkaria), Dutch and Spanish (CoNLL 2002) and German (GermEval 2014 outer). L: Local model. C: cluster model. $\Delta$: difference between them.}
  \label{tab:multi-learn-curv}
\end{table}

These results are mirrored by those obtained for the rest of the languages and datasets. Thus, Table \ref{tab:multi-learn-curv} displays, for each language, the F1 results of the baseline system and of the best \emph{cluster} models on top of the baseline.\footnote{The Basque dataset was far too small to train models with 1/16 of the data.} Overall, it confirms that our cluster-based models obtain state of the art results using just one half of the data. Furthermore, using just one quarter of the training data we are able to match results of other publicly available systems for every language, outperforming in some cases, such as Basque, much complex systems of classifiers exploiting linguistic specific rules and features (POS tags, lemmas, semantic information from WordNet, etc.). Considering that Basque is a low-resourced language, it is particularly relevant to be able to reduce as much as possible the amount of gold supervised data required to develop a competitive NERC system.

\subsection{Out-of-domain evaluations}\label{sec:out-doma-eval}

NERC systems are often used in out-of-domain settings, namely, to annotate data that greatly differs from the data from which the NERC models were learned. These differences can be of text genre and/or domain, but also because the assumptions of what constitutes a named entity might differ. It is therefore interesting to develop robust NERC systems across both domains and datasets. In this section we demonstrate that our approach, consisting of basic, general local features and the \emph{combination} and \emph{stacking} of clusters, produces robust NERC systems in three out-of-domain evaluation settings:

\begin{itemize}
\item Class disagreements: Named entities are assigned to different classes in training and test.
\item Different text genre: The text genre of training and test data differs.
\item Annotation guidelines: The gold annotation of the test data follows different guidelines from the training data. This is usually reflected in different named entity spans.
\end{itemize}

The datasets and languages chosen for these experiments are based on the availability of both previous results and publicly distributed NERC systems to facilitate direct comparison of our system with other approaches. Table \ref{tab:out-of-domain-datasets} specifies the datasets used for each out-of-domain setting and language. Details of each dataset can be found Table \ref{tab:datasets}.

\begin{table}[ht]\footnotesize
  \centering
  \begin{tabular}{cccccccccc}\hline
& \multicolumn{2}{c}{Class Disagreements} & \multicolumn{2}{c}{Text Genre} & \multicolumn{2}{c}{Annotation Guidelines} \\ \hline \hline
  & Train & Test & Train & Test & Train & Test \\ \hline
  en & CoNLL & MUC 7 & CoNLL & Wikigold & CoNLL/Ontonotes/MUC 7 & MEANTIME \\
  es & Ancora & CoNLL & - & - & Ancora/CoNLL & MEANTIME \\
  nl & SONAR-1 & CoNLL & - & - & SONAR-1 & MEANTIME \\ \hline
  \end{tabular}
  \caption{Testsets and languages for out-of-domain evaluations.}
  \label{tab:out-of-domain-datasets}
\end{table}

\subsubsection{Class Disagreements}\label{sec:class-disagreements}

MUC 7 annotates seven entity types, including four that are not included in CoNLL data: DATE, MONEY, NUMBER and TIME entities. Furthermore, CoNLL includes the MISC class, which was absent in MUC 7. This means that there are class disagreements in the gold standard annotation between the training and test datasets. In addition to the four CoNLL classes, SONAR-1 includes PRODUCT and EVENT whereas Ancora also annotates DATE and NUMBER. For example, consider the following sentence of the MUC 7 gold standard (example taken from \cite{ratinov_design_2009}):

\begin{center}
  ``...baloon, called the Virgin Global Challenger.''
\end{center}

The gold annotation in MUC 7 establishes that there is one named entity:

\begin{center}
  ``...baloon, called [ORG Virgin] Global Challenger.''
\end{center}

However, according to CoNLL 2003 guidelines, the entire name should be annotated like MISC:

\begin{center}
  ``...baloon, called [MISC Virgin Global Challenger].''
\end{center}

In this setting some adjustments are made to the NERC systems' output. Following previous work \citep{ratinov_design_2009}, every named entity that is not LOC, ORG, PER or MISC is labeled as `O'. Additionally for MUC 7 every MISC named entity is changed to `O'. For English we used the models reported in Section \ref{sec:english}. For Spanish and Dutch we trained our system with the Ancora and SONAR-1 corpora using the configurations described in Sections \ref{sec:spanish} and \ref{sec:dutch} respectively. Table \ref{tab:classout} compares our results with previous approaches: using MUC 7, \cite{turian-ratinov-bengio:2010:ACL} provide standard phrase results whereas \cite{ratinov_design_2009} score token based F1 results, namely, each token is considered a chunk, instead of considering multi-token spans too. For Spanish we use the Stanford NER Spanish model (2015-01-30 version) trained with Ancora. For Dutch we compare our SONAR-1 system with the companion system distributed with the SONAR-1 corpus \citep{desmet2014fine}. The results are summarized in Table \ref{tab:classout}.

\begin{table}[ht]\footnotesize
  \centering
  \begin{tabular}{lcclclc} \hline
    \multicolumn{3}{c}{muc7} & \multicolumn{2}{c}{es-conll} & \multicolumn{2}{c}{nl-conll} \\ \hline \hline
    Features & F1 & T-F1 & Features & F1 & Features & F1 \\ \hline
     Local & 71.89 & 74.53 & Local & 68.48 & Local & 57.53 \\
     BR & 80.06 & 82.19 & BP & 70.96 & BW & 61.53 \\
     CR600 + CW600 & 82.19 & 83.48 & CG400 + CW400 & 70.64 & CW400 & 61.59 \\
     W2VG200 & 77.12 & 78.86 & W2VG400 & 72.05 & W2VW100 & 61.67 \\
     en-91-18-conll03 (clusters) & 83.49 & 85.35 & es-clusters & \textbf{72.51} & nl-clusters & 62.90 \\
     en-91-18-conll03 + dict & \textbf{84.86} & \textbf{87.03} & es-clusters-dict & 71.88 & nl-clusters-dict & \textbf{64.16} \\ \hline
     Ratinov and Roth/Turian et al. & 84.15 & 86.15 & Stanford-ancora & 49.92 & sonar-nerd & 56.67 \\ \hline
  \end{tabular}
  \caption{Out-of-domain evaluation based on class disagreements. English models trained on CoNLL 2003; Spanish models trained with Ancora; Dutch models trained with SONAR-1. T-F1: token-based F1.}
  \label{tab:classout}
\end{table}

\subsubsection{Text Genre}\label{sec:text-genre}

In this setting the out-of-domain character is given by the differences in text genre between the English CoNLL 2003 set and the Wikigold corpus. We compare our system with English models trained on large amounts of silver-standard text (3.5M tokens) automatically created from the Wikipedia \citep{nothman_learning_2012}. They report results on Wikigold showing that they outperformed their own CoNLL 2003 gold-standard model by 10 points in F1 score. We compare their result with our \emph{best cluster} model in Table \ref{tab:wikigoldout}. While the results of our baseline model confirms theirs, our clustering model score is slightly higher. This result is interesting because it is arguably more simple to induce the clusters we use to train \emph{ixa-pipe-nerc} rather than create the silver standard training set from Wikipedia as described in \cite{nothman_learning_2012}.

\begin{table}[ht]\footnotesize
  \centering
  \begin{tabular}{lccc} \hline
    System & Precision & Recall & Phrase F1 \\ \hline \hline
     Local & 59.01 & 52.64 & 55.64 \\
     BR+BW & 62.58 & 57.76 & 60.06 \\
     CR600 + CW600 & 67.79 & 58.51 & 63.04 \\
     W2VG200 + W2VW400 & 64.32 & 58.61 & 61.39 \\
     best-cluster & \textbf{70.09} & 64.42 & \textbf{67.14} \\ \hline
     Stanford NER (distsim-conll03) & 64.40 & 61.89 & 63.12 \\
     en-wiki2 (Nothman et al. 2013) & 64.60 & \textbf{68.70} & 66.60 \\ \hline
  \end{tabular}
  \caption{Wikigold out-of-domain evaluation based on text genre.}
  \label{tab:wikigoldout}
\end{table}

\subsubsection{Annotation Guidelines}\label{sec:annot-guid}

In this section the objective is studying not so much the differences in textual genre as the influence of substantially different annotation standards.  We only use three classes (location, organization and person) to evaluate the best models presented for in-domain evaluations labeling `O' every entity which is not LOC, ORG or PER.

The text genre of MEANTIME is not that different from CoNLL data. However, differences in the gold standard annotation result in significant disagreements regarding the span of the named entities \citep{NWR-guidelines-final}. For example, the following issues are markedly different with respect to the training data we use for each language:

\begin{itemize}
    \item Different criteria to decide when a named entity is annotated: in the
        expression ``40 billion US air tanker contract'' the MEANTIME gold
        standard does not mark `US' as location, whereas in the training data this is systematically annotated.
    \item Mentions including the definite article within the name entity span: `the United States' versus `United States'.
    \item Longer extents containing common nouns: in the MEANTIME corpus there are many entities such as
        ``United States airframer Boeing'', which in this case is considered an
        organization, whereas in the training data this span will in general consists of two entities:
        `United States' as location and `Boeing' as organization.
    \item Common nouns modifying the proper name: `Spokeswoman Sandy Angers' is
        annotated as a named entity of type PER whereas in the training data used the
        span of the named entity would usually be `Sandy Angers'.
\end{itemize}

CoNLL NER phrase based evaluation punishes any bracketing error as both false positive and negative. Thus, these span-related disagreements make this setting extremely hard for models trained according to other annotation guidelines, as shown by Table \ref{tab:wikinewsmulti}. Our baseline models degrade around 40 F1 points and the cluster-based models around 35. Other systems' results worsen much more, especially for Spanish and Dutch. The token-based scores are in general better but the proportion in performance between systems across languages is similar.

\begin{table*}[ht]\footnotesize
  \centering
  \begin{tabular}{l*{12}{c}} \hline
    & \multicolumn{4}{c}{English} & \multicolumn{4}{c}{Spanish} & \multicolumn{4}{c}{Dutch} \\ \hline \hline
    & \multicolumn{2}{c}{Outer} & \multicolumn{2}{c}{Inner} & \multicolumn{2}{c}{Outer} & \multicolumn{2}{c}{Inner} & \multicolumn{2}{c}{Outer} & \multicolumn{2}{c}{Inner} \\ \hline \hline
    Features & F1 & T-F1 & F1 & T-F1 & F1 & T-F1 & F1 & T-F1 & F1 & T-F1 & F1 & T-F1 \\ \hline
    Local & 41.83 & 54.17 & 48.57 & 57.85 & 34.42 & 42.95 & 37.14 & 41.93 & 48.49 & 54.84 & 49.77 & 55.86 \\
    best-cluster & 54.04 & 65.96 & 63.72 & 71.13 & 56.78 & 62.55 & 59.77 & 63.04 & 59.94 & 66.03 & 60.27 & 65.42 \\
    best-overall & \textbf{55.48} & 67.36 & \textbf{64.95} & 71.98 & \textbf{58.94} & 65.63 & \textbf{62.14} & 65.54 & \textbf{63.40} & 70.68 & \textbf{63.93} & 70.24 \\ \hline
    Stanford NER & 53.14 & 64.62 & 62.45 & 69.76 & 46.42 & 54.40 & 47.48 & 54.27 & - & - & - & - \\
    Illinois NER & 53.24 & 65.68 & 62.72 & 71.04 & - & - & - & - & - & - & - & - \\
    Freeling 3.1 & - & - & - & - & 38.27 & 48.06 & 40.93 & 46.52 & - & - & - & - \\
    Sonar nerd & - & - & - & - & - & - & - & - & 48.60 & 53.60 & 48.44 & 52.79 \\ \hline
  \end{tabular}
  \caption{MEANTIME out-of-domain evaluation. English systems trained on CoNLL data. Dutch systems trained with SONAR-1. Stanford NER Spanish model is trained with Ancora (20150130 version) whereas ixa-pipe-nerc is trained with CoNLL data. T-F1: token-based F1. \emph{Local}: baseline system; \emph{best-clusters}: nl-clusters, es-cluster and en-best-cluster; \emph{best-overall}: best configuration previously presented for each language for the in-domain evaluations.}
  \label{tab:wikinewsmulti}
\end{table*}

As an additional experiment, we also tested the English model recommended by Stanford NER which is trained for three classes (LOC, PER, ORG) using a variety of public and (not identified) private corpora (referred to as Stanford NER 3 class (ALL) in Table \ref{tab:wikinewsout}). The results with respect to their CoNLL model improved by around 3 points in F1 score across named entity labels and evaluation types (phrase or token based). In view of these results, we experimented with multi-corpora training data added to our best CoNLL 2003 model (\emph{en-91-18}). Thus, we trained using three public training sets: MUC 7, CoNLL 2003 and Ontonotes 4.0. The local model with the three training sets (Local ALL) improved 12 and 17 points in F1 score across evaluations and entity types, outperforming our best model trained only with CoNLL 2003. Adding the clustering features gained between 2 and 5 points more surpassing the Stanford NER 3 class multi-corpora model in every evaluation. We believe that the main reason to explain these improvements is the variety and quantity of annotations provided by Ontonotes (1M word corpus), and to a lesser extent by MUC 7, which includes some spans containing common nouns and determiners making the model slightly more robust regarding the mention spans.

\begin{table}[ht]\footnotesize
  \centering
  \begin{tabular}{lcccc} \hline
    & \multicolumn{2}{c}{Outer NEs} & \multicolumn{2}{c}{Inner NEs} \\ \hline \hline
    Features & F1 & T-F1 & F1 & T-F1 \\ \hline
    Local (ALL) & 56.00 & 66.30 & 65.89 & 73.01 \\
    en-91-18-conll03 (ALL) & \textbf{62.09} & 70.98 & \textbf{70.90} & 77.18\\ \hline
    Stanford NER 3 class (ALL) & 57.14 & 70.22 & 66.96 & 76.01 \\ \hline
  \end{tabular}
  \caption{MEANTIME English multi-corpus out-of-domain evaluation.}
  \label{tab:wikinewsout}
\end{table}

\section{Discussion}\label{sec:disc-future-work}

Despite the simplicity of the \emph{ixa-pipe-nerc} approach, we report best results for English in 4 different datasets: for CoNLL 2003 and for the three English out-of-domain evaluations. For German we improve the results of the best system in the GermEval 2014 task and obtain comparable results to previous work in the CoNLL 2003 dataset using publicly available data. In Spanish we provide results on CoNLL 2002 and in two out-of-domain evaluations clearly outperforming previous best results. For Dutch we improve over previous results in CoNLL 2002 and SONAR-1 data and two out-of-domain evaluations. Finally, for Basque (Egunkaria) the improvements are considerable.

\paragraph{Simple and shallow features}

These results are obtained without linguistic or global features. Instead, injecting unigram knowledge from the \emph{combination} and \emph{stacking} of clusters allows to obtain a robust NERC system across languages, outperforming other, more complex \citep{ratinov_design_2009,turian-ratinov-bengio:2010:ACL,desmet2014fine,passos-kumar-mccallum:2014:W14-16} and language-specific systems. This is also the case for languages such as Basque or German, where the use of linguistic features (lemmas, POS tags, curated suffix lists and rules, etc.) has so far been pervasive \citep{alegria2006lessons,benikova7germeval,germaner}.

\paragraph{Minimal human intervention}

Each of the datasets used displays an idiosyncratic annotation and genre. This is even the case for the NER tasks organized at CoNLL 2002 and 2003: ``For instance, Spanish marks no lowercase adjectival nationalities and includes 192 instances where surrounding quotes are included in the entity annotation; Dutch has as PER the initials of photographers; and English has lots of financial and sports data in tables'' \citep{nothman_learning_2012}.

In despite of this, our best in-domain results were obtained using the same set of features for all seven evaluations, which included trigrams and character n-grams. The only variable across datasets and languages was the number of classes of the clustering lexicons used.

However, the in-domain results also manifest that trigrams and character n-grams can be omitted for languages without declension cases or repeated suffixes in the named entities (e.g., Dutch, English and Spanish) without it being too detrimental. In fact, we started experimenting without trigrams and character n-grams for Dutch, English and Spanish. When we added them to the best model of each language (e.g, charngram 1:6 en-91-18 in Table \ref{tab:englishconll03}), the in-domain results improved or remained quite similar but at the cost of making the models less robust in the out-of-domain evaluations. In contrast, trigrams and character n-grams were highly beneficial in both in-domain and out-of-domain settings for Basque and German.

Our take on this issue is that trigram and character n-gram features would only be required to address inflected named entities (in Basque) or to learn repeated suffixes appearing in named entities and to tackle sparsity (in both Basque and German). For example, Table \ref{tab:ekuadorko} shows the utility of character n-gram features capturing the Basque locative declension case \emph{-ko}, which is repeated for many location entities in the training data.

The emphasis on clustering features for good performance (as opposed to local features) produces an easily exportable and robust system for both in-domain and out-of-domain evaluations and across languages. It is therefore crucial, for competitive performance, to understand which clustering methods and corpora use as well as how to combine them effectively.

\paragraph{Choosing the right corpus and clustering method}

Contrary to previous suggestions that the larger the number of classes and the corpus used to induced the clusters the better \citep{turian-ratinov-bengio:2010:ACL}, our results provide a number of interesting pointers to choose the appropriate type of corpus and clustering method required for optimal performance.

With respect to Brown clusters, all our results are better when we induce 1000 classes. We systematically tried for every language and data source with less (320) and more (3200) classes without performance improvement. Moreover, in every evaluation setting the best results with Brown clusters were obtained when a corpus relatively closed in-domain, genre and date was used, even if significantly smaller. This is especially clear for Basque, English and Spanish where the best Brown clusters were induced over the smallest corpora (Egunkaria, Reuters RCV1 and El Periodico, respectively).

In contrast, results show that Word2vec clusters, unlike Brown, always benefit from very large amounts of data, regardless of domain or temporal issues. Our experiments also suggest that Clark clusters seem to behave more robustly than Brown clusters with respect to the size and type of text, performing well with large unrelated and smaller domain-specific corpora. For best performance, \cite{clark2003combining} recommends that the proportion of clusters $k$ with respect to the source data should be of $k^3 \approx n$ where $n$ is the number of words in the corpus. Instead, we systematically induce, for every corpus, Clark clusters in the range of 100-600 classes, because preliminary experiments proved that over 600 classes, even if the proportion proposed by Clark holds, performance starts to deteriorate\footnote{Note that we did not experiment with any of the clustering algorithms' parameters, vector dimensions, etc, just with the number of classes.}. Following this, we are now in better position to address the questions posed by \cite{turian-ratinov-bengio:2010:ACL}:

\begin{itemize}
\item Brown clusters benefit from source data closely related to the testset, even if small in size.
\item Clark clusters behave robustly with respect to the size or type of data sources from which they are induced.
\item For Word2vec clusters size is the most important factor: the larger the corpus the better.
\end{itemize}

Should we prefer certain word features? While Clark features seem to obtain the best results overall, our work provides a very simple method of effectively combining them, depending on the data sources we have available.

\paragraph{Combination and Stacking}

We use three different data sources, namely, Wikipedia, Egunkaria and Berria (see Table \ref{tab:unlabeledcorpora} for the list of unlabeled corpora used) for the Basque experiments. In order to understand better our approach, we annotated the Basque testset with every model in Table \ref{tab:basqueresults} and manually inspected their output. The following two examples illustrate how our approach works:

\begin{itemize}
\item \emph{Ekuadorko}: In addition to the \emph{eu-cluster} model, the Brown model (BE), and the Clark Wikipedia model (CW200) provide the correct annotation. The assigned clusters in BE and CW200 clustering lexicons clearly consist of locations. For example, the 176 cluster of CW200 contains \emph{Gasteizko}, \emph{Arabako}, \emph{Espainiako}, etc., which, unlike \emph{Ekuadorko}, do occur in the training set.\footnote{English: in Vitoria-Gasteiz, in Alava, in Spain.} Most interestingly, while \emph{Gasteizko} is only labeled as location in the training set, both \emph{Arabako} and \emph{Espainiako} are labeled as, depending on the context, organization or location.
\item \emph{Ameriketara} (to America): Only the Brown model (and the \emph{eu-cluster}) correctly labels it as a location. In this case, the \emph{011110011100} Brown path clusters \emph{Ameriketara} with other locations such as \emph{Baionara} and \emph{Espainiara}, among others, which, unlike \emph{Ameriketara}, are contained in the training set.\footnote{To Bayonne, to Spain.}
\end{itemize}

The same phenomenon can be observed for languages quite different from Basque such as English or Spanish. For example, the named entity `Uzbekistan' is not present in the English CoNLL 2003 training data, whereas in the test set can be found four times, all of them locations. The local model annotates all four occurrences as organization (see Table \ref{tab:englishconll03} for references to models). The Brown and the Word2vec models, two as locations and two as organizations, because the cluster companions of `Uzbekistan' are of mixed nature. Finally, the Clark model (CW600) does correctly annotate them as locations (also \emph{best-cluster}, \emph{en-91-18}, \emph{en-91-36} $\ldots$): `Uzbekistan' is placed in the 145 cluster, which contains mostly locations contained in the training data (Spain, Ireland, etc.).

Previous approaches to NERC combining clusters or word embeddings have obtained mixed results \citep{turian-ratinov-bengio:2010:ACL}. Up until now best results have been based on rather complex systems which also used one type clustering or embedding feature \citep{passos-kumar-mccallum:2014:W14-16,ratinov_design_2009,faruqui_training_2010,benikova7germeval}. In other sequence labeling tasks, \cite{biemann2009unsupervised} reports a slight improvement (from 97.33 to 97.43 word accuracy) in POS tagging combining two types of clustering methods (one of them was \cite{clark2003combining}) for German.

Our system displays two important differences with respect to previous approaches. First, the differences between our baseline system and the, for example, Clark features are much larger than in previous work (with the exception of \cite{faruqui_training_2010}), ranging from 2.2 and 5.5 points in F1 measure across the in-domain evaluations to 2-8 points for out-of-domain results. For example, our English CoNLL 2003 single clustering models are similar to the best CoNLL 2003 model distributed by the Stanford NER. If we consider the combined clustering models, the differences over the baseline increase to 5-10 points of F1 measure for in-domain evaluations and between 4-22 in out-of-domain settings.

Second, our combination of clustering features significantly increases the performance over the models using only one type of clustering feature. The improvements range over 2 to 6 points in F1 measure for in-domain and out-of-domain results.

In our opinion, these results are quite interesting as previous experiments combining features of different word representations for NERC \citep{turian-ratinov-bengio:2010:ACL}, while increasing the overall result, did not improve over the state of the art at the time \citep{ratinov_design_2009}. The results also show that leaning heavily on the clustering features (instead of specific feature tuning) for performance proves very beneficial in out-of-domain settings.

\paragraph{Robust in out-of-domain settings}

The results of the eight out-of-domain evaluations undertaken suggest that differences regarding named entities spans as described in Section \ref{sec:annot-guid} are harder to overcome than disagreements in text genre (e.g. Section \ref{sec:text-genre}) or entity type (Section \ref{sec:class-disagreements}). Thus, our method using multiple clustering sources allow to overcome better any differences in named entity type or text genre. However, and even though our system obtains state of the art results in every evaluation, trying to adapt to differences in named entity shape and span proves to be a much more difficult task, hence the comparatively lower results obtained in the MEANTIME evaluations.

\paragraph{Robust reducing training data}

\cite{koo-carreras-collins:2008:ACLMain} present learning curves showing the increase in performance when using Brown clusters for dependency parsing whereas \cite{biemann2009unsupervised} provides learning curves to measure the impact of clusters for NERC and chunking. Inspired by those two previous works we measured the performance when training data is reduced. Unlike these two approaches, the differences between adding clusters or not to our system with less training data is huge. Table \ref{tab:multi-learn-curv} shows that differences adding the clustering features with half the data is around 8 points in F1 score (for Spanish the difference is 3.42 F1).

Another common point of our clustering features with \cite{koo-carreras-collins:2008:ACLMain} is that when gold training data is reduced, the system still obtains competitive results with respect to previous approaches or publicly available systems using only a fraction (half or a quarter) of the data. If we consider the ability of our system to be competitive with substantial reductions in gold training data plus the fact that no linguistic motivated features are required, we are providing a system which is much cheaper to train for new languages and/or domains contributing therefore to alleviate the dependency on gold training data to obtain good performing NERC systems for new languages, domains or genres.

\section{Conclusion and Future Work}\label{sec:conclusion}

We have shown how to develop robust NERC systems across languages and datasets with minimal human intervention, even for languages with inflected named entities. This is based on adequately combining word representation features on top of shallow and general local features. Crucially, we have empirically demonstrate how to effectively combine various types of simple word representation features depending on the source data available. This has resulted in a clear methodology for using the three types of clustering features which produces very competitive results in both in-domain and out-of-domain settings.

Thus, despite the relative simplicity of our approach, we report state of the art results for Dutch, English, German, Spanish and Basque in seven in-domain evaluations.

We also outperform previous work in eight out-of-domain evaluations, showing that our clustering features improve the robustness of NERC systems across datasets. Finally, we have measured how much our system's performance degrades when the amount of supervised data is drastically cut. The results show our models are still very competitive even when reducing the supervised data by half or more. This, together with the lack of linguistic features, facilitates the easy and fast development of NERC systems for new domains or languages.

In future work we would like to explore more the various types of domain adaptation required for robust performance across text genres and domains, perhaps including micro-blog and noisy text such as tweets. Furthermore, we are also planning to adapt our techniques to other sequence labeling problems such as Opinion Target Extraction \citep{pontiki2014semeval,pontiki-EtAl:2015:SemEval} and Super Sense tagging \citep{ciaramita2006broad}.

\section*{Acknowledgments}

We would like to thank the anonymous reviewers for their comments to improve this paper. We would also like to thank Sebastian Pad\'o for his help training the Clark clusters. This work has been supported by the European projects NewsReader, EC/FP7/316404 and QTLeap - EC/FP7/610516, and by the Spanish Ministry for Science and Innovation (MICINN) SKATER, Grant No. TIN2012-38584-C06-01 and TUNER, TIN2015-65308-C5-1-R.

\end{document}